\documentclass[conference]{IEEEtran}
\usepackage[latin1]{inputenc}
\usepackage{makeidx}          
\usepackage{graphics,graphicx}          
\usepackage{amssymb,amsmath,amsthm}  
\usepackage{subfigure}
\usepackage{textcomp}
\usepackage[margin=19mm]{geometry}
\usepackage{algorithm}
\usepackage{algorithmic}
\usepackage{float}
\usepackage{color}
\usepackage{verbatim}
\usepackage{enumerate}
\usepackage{url}

\IEEEoverridecommandlockouts

\title{Towards Explainable Deep Neural Networks (xDNN) 
\\ {\large Preprint submitted to Neural Networks Journal }
\thanks{Plamen Angelov, and Eduardo Soares are with the School of Computing and Communications, Lancaster University, Lancaster, LA1 4WA, UK.  E-mails: p.angelov@lancaster.ac.uk; e.almeidasoares@lancaster.ac.uk.}}

\author{Plamen Angelov, Eduardo Soares}

\begin{document}

\maketitle

\begin{abstract}
In this paper, we propose an elegant solution that is directly addressing the bottlenecks of the traditional deep learning approaches and offers a clearly explainable internal architecture that can outperform the existing methods, requires very little computational resources (no need for GPUs) and short training times (in the order of seconds). The proposed approach, xDNN is using prototypes. Prototypes are actual training data samples (images), which are local peaks of the empirical data distribution called \textit{typicality} as well as of the data density. This generative model is identified in a closed form and equates to the pdf but is derived automatically and entirely from the training data with no user- or problem-specific thresholds, parameters or intervention. The proposed xDNN offers a new deep learning architecture that combines reasoning and learning in a synergy. It is non-iterative and non-parametric, which explains its efficiency   in terms of time and computational resources. From the user perspective, the proposed approach is clearly understandable to human users. We tested it on some well-known benchmark data sets such as iRoads and Caltech-256. xDNN outperforms the other methods including deep learning in terms of accuracy, time to train and offers a clearly explainable classifier. In fact, the result on the very hard Caltech-256 problem (which has 257 classes) represents a world record \cite{he2015spatial}.

\end{abstract}

\section{Introduction}

Deep learning has demonstrated ability to achieve highly accurate results in different application domains such as speech recognition \cite{xiong2018microsoft}, image recognition \cite{he2016deep}, and language translation \cite{lecun2015deep} and other complex problems \cite{goodfellow2016deep}. It attracted the attention of media and the wider public \cite{sejnowski2018deep}. It has also proven to be very valuable and efficient in automating the usually laborious and sometimes controversial pre-processing stage of feature extraction. The main criticism towards deep learning is usually related to its `black-box' nature and requirements for huge amount of labeled data, computational resources (GPU accelerators as a standard), long times (hours) of training, high power and energy requirements \cite{rudin2019stop}. Indeed, a traditional deep learning (e.g. convolutional neural network) algorithm involves hundreds of millions of weights/coefficients/parameters that require iterative optimization procedures. In addition, these hundreds of millions of parameters are abstract and detached from the physical nature of the problem being modelled. However, the automated way to extract them is very attractive in high throughput applications of complex problems like image processing where the human expertise may simply be not available or very expensive. 

Feature extraction is an important pre-processing stage, which defines the data space and may influence the level of accuracy the end result provides. Therefore, we consider this very useful property of the traditional deep learning and step on it combined with another important recent result in the deep learning domain, namely, the transfer learning. This concept postulates that knowledge in the form of a model architecture learned in one context can be re-used and useful in another context \cite{hu2015deep}. Transfer learning helps to considerably reduce the amount of time used for training. Moreover, it also may help to improve the accuracy of the models \cite{zhuang2015supervised}.

Stepping on the two main achievements of the deep learning - top accuracy combined with an automatic approach for feature extraction for complex problems, such as image classification, we try to address its deficiencies such as the lack of explainability \cite{rudin2019stop}, computational burden, power and energy resources required, ability to self-adapt and evolve \cite{soares2019novelty}. Interpretability and explainability are extremely important for high stake applications, such as autonomous cars, medical or court decisions, etc. For example, it is extremely important to know the reasons why a car took some action, especially if this car is involved in an accident \cite{doshi2017towards}. 

\begin{figure}[h]
	\begin{center}
		{\includegraphics[scale=1]{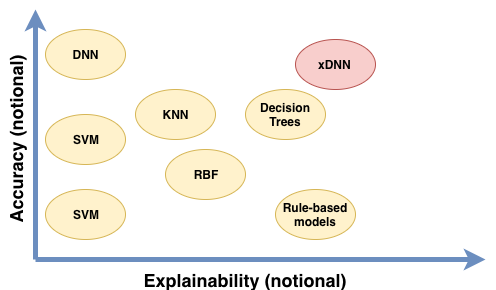}}
		\caption{Trade-off between accuracy and explainability.  } \label{Fig1}
	\end{center}
\end{figure}

The state-of-the-art classifiers offer a choice between higher explainability for the price of lower accuracy or vice versa (Figure \ref{Fig1}). Before deep learning \cite{schmidhuber2015deep}, machine-learning and pattern-recognition required substantial domain expertise to model a feature extractor that could transform the raw data into a feature vector which defines the data space within which the learning subsystem could detect or classify data patterns \cite{lecun2015deep}. Deep learning offers new way to extract abstract features automatically. Moreover, pre-trained structures can be reused for different tasks through the transfer learning technique \cite{hu2015deep}. Transfer learning helps to considerably reduce the amount of time used for training, moreover, it also may helps to improve the accuracy of the models \cite{zhuang2015supervised}. In this paper, we propose a new approach, xDNN that offers both, high level of explainability combined with the top accuracy.

The proposed approach, xDNN offers a new deep learning architecture that combines reasoning and learning in a synergy. It is based on prototypes and the data density \cite{angelov2019empirical} as well as \textit{typicality} - an empirically derived pdf \cite{angelov2017generalized}. It is non-iterative and non-parametric, which explains its efficiency in terms of time and computational resources. From the user perspective, the proposed approach is clearly understandable to human users. We tested it on some well-known benchmark data sets such as iRoads \cite{rezaei2013vehicle} and Caltech-256 \cite{griffin2007caltech} and xDNN outperforms the other methods including deep learning in terms of accuracy, time to train, moreover, offers a clearly explainable classifier. In fact, the result on the very hard Caltech-256 problem (which has 257 classes) represents a world record \cite{he2015spatial}. 

The remainder of this paper  is  organized  as  follows:  The next section introduces the proposed explainable deep learning approach. The experimental data employed in the analysis and results are presented in the results section. Discussion is presented in the last section of this paper.

\section{Explainable Deep Neural Network}

\subsection{Architecture and Training of the proposed xDNN}
The proposed explainable deep neural network (xDNN) classifier is formed of several layers with a very clear semantic and functional meaning. In addition to the internal clarity and transparency it also offers a very clear from the user point of view set of prototype-based $IF...THEN$ rules. Prototypes are selected data samples (images) that the user can easily view, understand and appreciate the similarity to other validation images. xDNN offers a synergy between the statistical learning and reasoning bringing both together. In most of the other approaches there is a dichotomy and preference of one over the other. We advocate and demonstrate that both, learning and reasoning can work together in a synergy and produce very impressive results. Indeed, the proposed xDNN method outperforms all published results \cite{rezaei2013vehicle,he2015spatial,angelov2018deep} in terms of accuracy. Moreover, in terms of time for training, computational simplicity, low power and energy required it is also far ahead.
The proposed approach can be described as a feedforward neural network which has an incremental learning algorithm that autonomously self-develops and evolves its structure adding new prototypes to reflect the possibly changing (dynamically evolving) data pattern \cite{soares2019novelty}. As shown in Figure \ref{fig3}, xDNN is composed of the following layers-- 

\begin{enumerate}
    \item Features descriptor layer; 
    \item Density layer; 
    \item Typicality layer;
    \item Prototypes layer; 
    \item \textit{MegaClouds} layer; 
\end{enumerate}

\begin{figure*}
  \centering
  \includegraphics[width=1\linewidth]{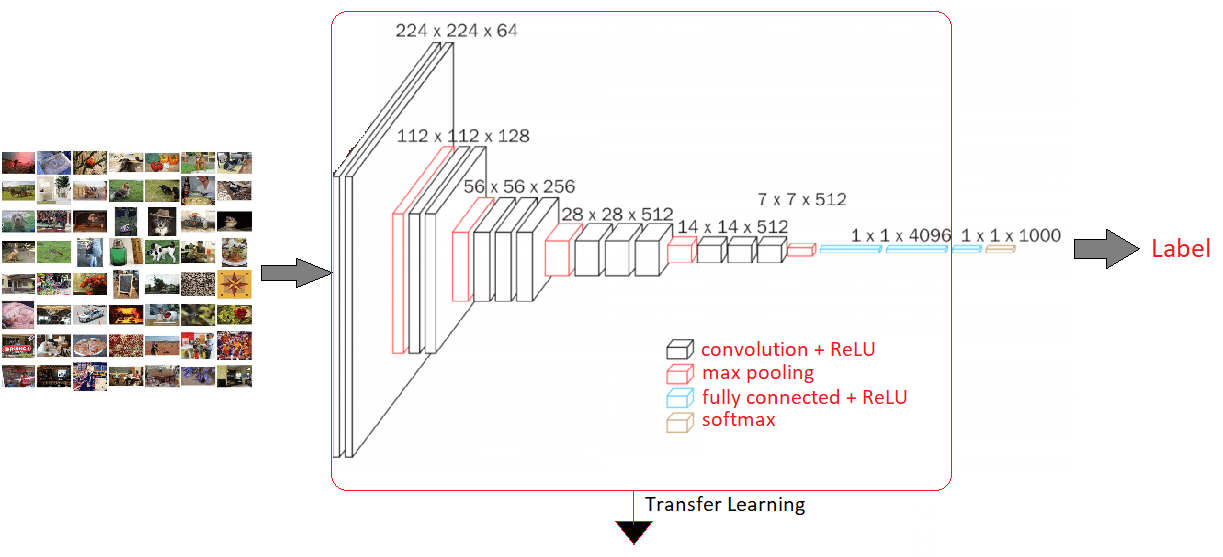}
  \caption{Pre-training a traditional deep neural network (weights of the network are being optimized/trained). Using the transfer learning concept this architecture with the weights are used as feature extractor (the last fully connected layer is considered as a feature vector). Adapted from \cite{simonyan2014very}.}
  \label{fig2}
\end{figure*}%
\begin{figure*}
  \centering
  \includegraphics[width=1\linewidth]{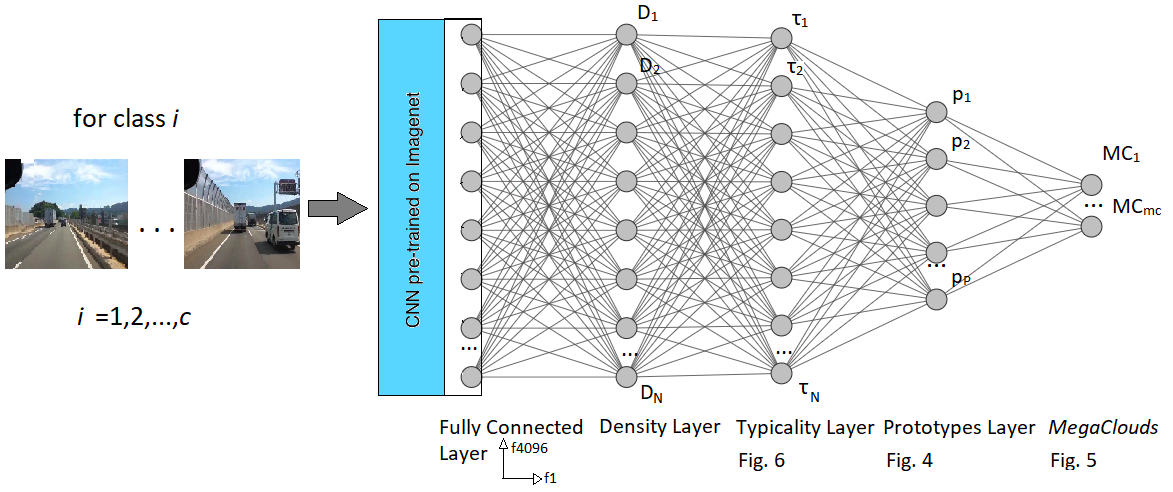}
  \caption{xDNN training architecture (per class).}
  \label{fig3}
\end{figure*}

\begin{enumerate}

\item \textbf{Features descriptor layer}: (Defines the data space)

The Feature Descriptor Layer is the first phase of the proposed xDNN method. This layer is in charge of extracting global features vector from the images. This
first layer can be formed by more traditional `handcrafted' methods such as GIST \cite{solmaz2013classifying} or HoG \cite{mizuno2012architectural}. Alternatively, it can be formed by the fully connected layer (FCL) of the pre-trained convolutional neural network approaches such as AlexNet \cite{krizhevsky2012imagenet}, VGG--VD--16 \cite{simonyan2014very}, and Inception \cite{szegedy2015going}, residual neural networks such as Resnet \cite{he2016deep} or Inception-Resnet \cite{szegedy2017inception}, etc. Using pre-trained deep neural network approach allows automatic extraction of more abstract and discriminative high-level features. 
In this paper, pre-trained VGG--VD--16 DCNN is employed for feature extraction. According to \cite{ren2016object},  VGG--VD--16 has a simple structure and it can achieve a better performance in comparison with other pre-trained deep neural networks. The first fully connected layer from VGG--VD--16 provides a $1 \times 4096$ dimensional vector. 

a)	The values are then standardized using the following equation (\ref{std}):

\begin{eqnarray}
\widehat{x}_{i,j} = \frac{x_{i,j} - \mu(x_{i,j})}{\sigma(x_{i,j})}
\label{std}
\end{eqnarray}

\noindent where $\widehat{x}$ denotes a standardized features vector $x$ of the image $I$ ($x$ are the values provided by the FCL), $i=1,2,...,N$ denotes the time stamp or the ID of the image, $j=1,2,...,n$ refers to the number of features of the given $x$ in our case $n=4096$.

b)	The standardized values are normalised to bring them to the range [0;1]:

\begin{eqnarray}
\Bar{x}_{i,j} = \frac{\widehat{x}_{i,j} - \min\limits_{i}(\widehat{x}_{i,j})}{\max\limits_{i}(\widehat{x}_{i,j})-\min\limits_{i}(\widehat{x}_{i,j})}
\label{Si}
\end{eqnarray}

\noindent where $\Bar{x}$ denotes the normalized value of the features vector. For clarity in the rest of the paper we will use $x$ instead of $\Bar{x}$.

\textbf{Initialization}:

Meta-parameters for the xDNN are initialized with the first observed data sample (image). The proposed algorithm works per class; therefore, all the calculations are done for each class separately. 

\begin{equation}
P \leftarrow 1;~~~\mu \leftarrow {x}_{i};
\end{equation}
where $\mu$ denotes the global mean of data samples of the given class. $P$ is the total number of the identified prototypes from the observed data samples (images).

Each class $C$ is initialized by the first data sample of that class:
\begin{equation}
\begin{split}
\mathrm{C}_{1} \leftarrow x_{1};~~~p_{1} \leftarrow x_{1};\\
Support_{1} \leftarrow 1;~~~r_{1}\leftarrow r^*;~~~\hat{I}_{1}\leftarrow I_{1}
\end{split}
\end{equation}

where, $p_{1}$ is the vector of features that describe the prototype $\hat{I}$ of the $C_{1}$; $\hat{I}$ is the identified prototype; $Support_{1}$ is the corresponding support (number of members) associated with this prototype; $r_{1}$ is the corresponding radius of the area of influence of $C_{1}$.

In this paper, we use $r^*=\sqrt{2-2cos(30^o)}$ same as \cite{angelov2019empirical}; the rationale is that two vectors for which the angle between them is less than $\pi/6$ or $30^o$ are pointing in close/similar directions $d$. That is, we consider that two feature vectors can be considered to be similar if the angle between them is smaller than 30 degrees. Note that $r^*$ is data derived, not a problem- or user- specific parameter. In fact, it can be defined without \textit{prior} knowledge of the specific problem or data through the following equation (\ref{eqcos}).

\begin{eqnarray}
d(x_i,p_i)=\left \| \frac{x_i}{\left \| x_i \right \|} - \frac{p_i}{\left \| p_i \right \|} \right \|. 
\label{eqcos}
\end{eqnarray}

\item \textbf{Density layer}: 

The density layer defines the mutual proximity of the images in the data space defined by the features from the previous layer. The data density, if use Euclidean form of distance, has a Cauchy form (\ref{eqS}) \cite{angelov2019empirical}:

\begin{eqnarray}
D(x_i) = \frac{1}{1+\frac{||x_i-\mu_N||^2}{\sigma^2_N}}, 
\label{eqS}
\end{eqnarray}

\noindent where $D$ is the density, $\mu$ is the global mean, and $\sigma$ is the variance. The reason it is Cauchy is not arbitrary \cite{angelov2019empirical}. It can be demonstrated theoretically that if Euclidean or Mahalanobis type of distances in the feature space are considered, the data density reduces to Cauchy type as referred in equation (\ref{eqS}). Density can also be updated online \cite{angelov2012autonomous}:

\begin{equation}
D(x_i)=\frac{1}{1+||x_i-\mu_i||^ 2 + \sum_i -||\mu_i||^2}.
\label{eqy}
\end{equation}

\noindent where $\mu_i$ and the scalar product, $\sum_i$ can be updated recursively as follows: 

\begin{equation}
\mu_i=\frac{i-1}{i}\mu_{i-1}+\frac{1}{i}x_i,
\label{eqy}
\end{equation}

\begin{equation}
\sum_i=\frac{i-1}{i}\sum_{i-1}+\frac{1}{i}||x_i||^ 2 ~ ~ \sum_1=||x_1||^ 2.
\label{eqy}
\end{equation}

Data samples (images) that are closer to the global mean have higher density values. Therefore, the value of the data density indicates how strongly a particular data sample is influenced by other data samples in the data space due to their mutual proximity.

\item \textbf{Typicality layer}: 

\textit{Typicality} is is an empirically derived form of probability distribution function (pdf). \textit{Typicality} $\tau$ is given by the equation (\ref{tau}).
The value of $\tau$ even at the point $x=p_i$ is much less than 1; the integral of $\int_{-\infty }^{\infty}\tau dx =1$ \cite{angelov2019empirical}.
\begin{equation}
\tau(x_i)= \frac{\sum_{i=1}^c Support_i D(x_i)}{\sum_{i=1}^c Support_i \int_{-\infty}^\infty D(x_i)dx}
\label{tau}
\end{equation}


\item \textbf{Prototypes layer}: 

The prototypes identification layer is the core of the proposed xDNN classifier. This layer is responsible to provide the clearly explainable model. The xDNN classifier is free from \textit{prior} assumptions about the data distribution type, as well as the random or deterministic nature of the data. In contrast, it extracts the actual distribution empirically form the data samples (images) bottom up \cite{angelov2019empirical}. The prototypes are independent from each other. Therefore, one can change the structure by adding a new prototype without influencing the other already existing prototypes. In other words, the proposed xDNN is highly parallelizable and suitable for evolving form of application where new prototypes may be added (if the data pattern requires this). The proposed xDNN method is trained per class forming a set of prototypes per class. Therefore, all the calculations are done for each class separately. 
Prototypes are the local peaks of the data density (and \textit{typicality}) identified in the previous layers/ stages of the algorithm from the images of the corresponding class based on their feature vectors. The prototypes can be used to form linguistic logical $IF ... THEN$ rules of the following form:

\vspace{8pt}

\noindent $R_c $:  IF $(I \sim \hat{I}_P)$ THEN $(class$ $ c)$

\noindent where $ \sim $ stands for similarity, it also can be seen as a fuzzy degree of membership; $p$ is the identified prototype; $P$ is the number of identified prototypes; $c$ is the class $c=1,2,...,C$, $I$ denotes an image.

One rule per prototype can be formed. All rules per class can be combined together using logical OR, also known as disjunction or S-norm:

\vspace{8pt}

\noindent $R_c$: IF $( I \sim \hat{I}_1)$ OR $( I \sim \hat{I}_2)$ OR ... OR $( I \sim \hat{I}_P)$ THEN $(class$ $c)$

\vspace{8pt}

Figure \ref{FigVoronoi} illustrates the area of influence of the identified prototypes. These areas around the identified prototypes are called \textit{data clouds} \cite{angelov2019empirical}. Thus, each prototype defines a \textit{data cloud}.

\begin{figure}[H]
	\begin{center}
		{\includegraphics[scale=0.33]{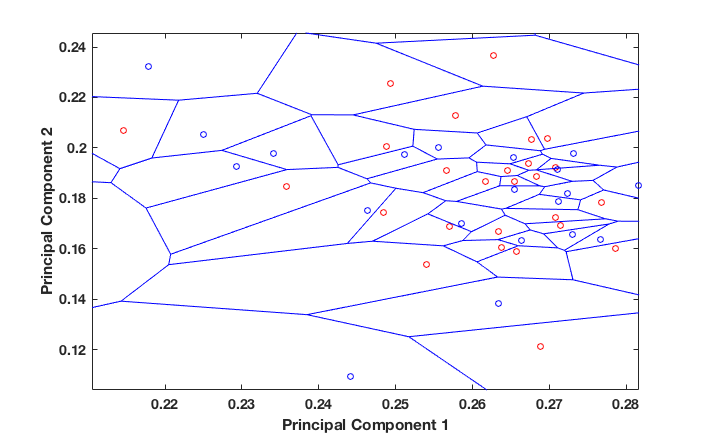}}
		\caption{Identified prototypes -- Voronoi Tesselation.  } \label{FigVoronoi}
	\end{center}
\end{figure}
\vspace{8pt}

 We call all data points associated with a prototype \textit{data clouds}, because their shape is not regular (e.g., hyper-spherical, hyper-ellipsoidal, etc.) and the prototype is not necessarily the statistical and geometric mean , but actual image \cite{angelov2019empirical}. The algorithm absorbs the new data samples one by one by assigning then to the nearest (in the feature space) prototype:

\begin{equation}
j^*=\operatorname*{argmin}_{j=1,2,...,P}(||x_{i}-p_{j}||^2)
\label{nearest}
\end{equation}

In case, the following condition \cite{angelov2019empirical} is met:
\begin{equation}
\begin{split}
\textit{\text{IF }}(D(x_{i})\geq\max_{j=1,2,...,P}D(p_j))~~\\
\textit{\text{OR }}~~(D(x_{i})\leq \min_{j=1,2,...,P}D(p_j))\\
\textit{\text{ THEN }} (add~a~new~data~cloud ~(P \leftarrow P+1))\label{if}
\end{split}
\end{equation}

It means that $x_{i}$ is out of the influence area of $p_j$. Therefore, the vector of features $x_{i}$ becomes a new prototype of a new \textit{data cloud} with meta-parameters initialized by equation (\ref{eqIN}).
 Add a new \textit{data cloud}:
\begin{equation}
\begin{split}
P \leftarrow P+1;~~~\mathrm{C}_{P} \leftarrow x_{i};
p_{P} \leftarrow I_{i};~~~Support_{P} \leftarrow 1;\\
~r_{P}\leftarrow r_o; \hat{I}_P \leftarrow I_{i};
\end{split}
\label{eqIN}
\end{equation}

Otherwise, \textit{data cloud} parameters are updated online by equation (\ref{update}). It has to be stressed that all calculations per \textit{data cloud} are performed on the basis of data points associated with a certain \textit{data cloud} only (i. e. locally, not globally, on the basis of all data points).

\begin{equation}
\begin{split}
C_{j^*} \leftarrow C_{j^*} + 1; ~~\\
p_{j^*} \leftarrow \frac{Support_{j^*}}{Support_{j^*}+1}p_{j^*}+\frac{Support_{j^*}}{Support_{j^*}+1}x_{i}; ~~ \\
Support_{j^*} \leftarrow Support_{j^*}+1; ~~ \\
r_{j^*}^2 \leftarrow \frac{r_{j^*}^2 +(1-||p_{j^*}||^2)}{2}.
\end{split}
\label{update}
\end{equation}

The xDNN learning procedure can be summarized by the following algorithm.
\vspace{6pt}
\hrule
\vspace{6pt}
\textbf{xDNN: Learning Procedure}
\vspace{3pt}
\hrule
\vspace{4pt}
\begin{algorithmic}[1]

    \STATE Read the first feature vector sample $x_i$ representing the image $I_i$ of the class $c$;
	\STATE Set $i \leftarrow 1; n \leftarrow 1; P_1 \leftarrow 1; p_1 \leftarrow x_i;  \mu  \leftarrow  x_1; Support \leftarrow   1; r_1 \leftarrow  r_0; \hat{I_1} \leftarrow I_1$;
	\STATE \textbf{FOR} $i$ = 2, ...
	\STATE ~~ Read $x_i$;
	\STATE ~ Calculate $D(x_i)$ and $D(p_j)$ $(j=1,2,...,P)$ according to equation (\ref{eqy});
	\STATE ~~ \textbf{IF} Equation (\ref{if})  holds
	\STATE ~~~~ Create rule according to Equation (\ref{eqIN});
	\STATE ~~ \textbf{ELSE}
	\STATE ~~~~ Search for $p_j$ according to Equation (\ref{nearest});
	\STATE ~~~~ Update rule according to Equation (\ref{update});
	\STATE ~~ \textbf{END}
	\STATE \textbf{END}	
\end{algorithmic}
\vspace{4pt}
\hrule
\vspace{6pt}

\item \textbf{\textit{MegaClouds} layer}: 

In the \textit{MegaClouds} layer the \textit{clouds} formed by the prototypes in the previous layer are merged if the neighbouring prototypes have the same class label. In other words, they are merged if they belong to the same class. \textit{MegaClouds} are used to facilitate the human interpretability. Figure \ref{FigMegaClouds} illustrates the formation of the \textit{MegaClouds}.

\begin{figure}[H]
	\begin{center}
		{\includegraphics[scale=0.33]{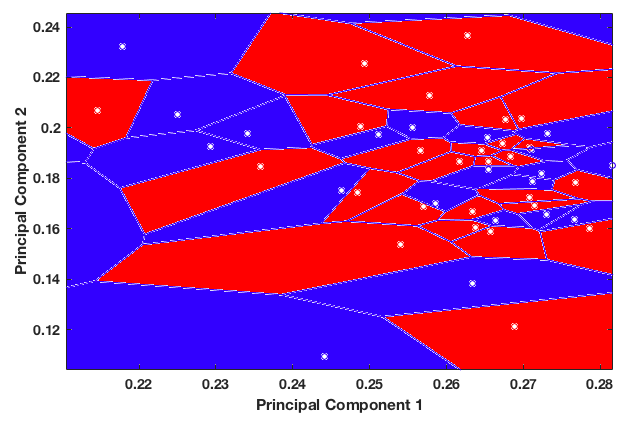}}
		\caption{\textit{MegaClouds} -- Voronoi Tesselation.  } \label{FigMegaClouds}
	\end{center}
\end{figure}
\vspace{8pt}

Rules in the \textit{MegaClouds} layer have the following format:

\vspace{8pt}

\noindent $R_c$: IF $( x \sim MC_1)$ OR $( x \sim MC_2)$ OR ... OR $( x \sim MC_{mc})$ THEN $(class$ $c)$

\vspace{8pt}

\noindent where $MC$ are the \textit{MegaClouds}, or the areas formed from the merging of the \textit{clouds}, and $mc$ is the number of identified \textit{MegaClouds}. Multimodal \textit{typicality}, $\tau$, can  also be used to illustrate the \textit{MegaClouds} as illustrated by Figure \ref{Fig6}.

\begin{figure}[H]
	\begin{center}
		{\includegraphics[scale=0.35]{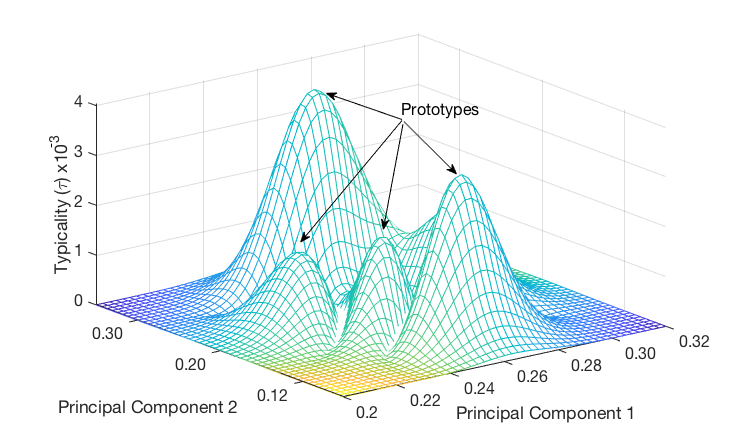}}
		\caption{\textit{Typicality} for the iRoads dataset. } \label{Fig6}
	\end{center}
\end{figure}

\end{enumerate}

\subsection{Architecture and Validation of the proposed xDNN}

\begin{figure*}[h]
	\begin{center}
		{\includegraphics[scale=0.33]{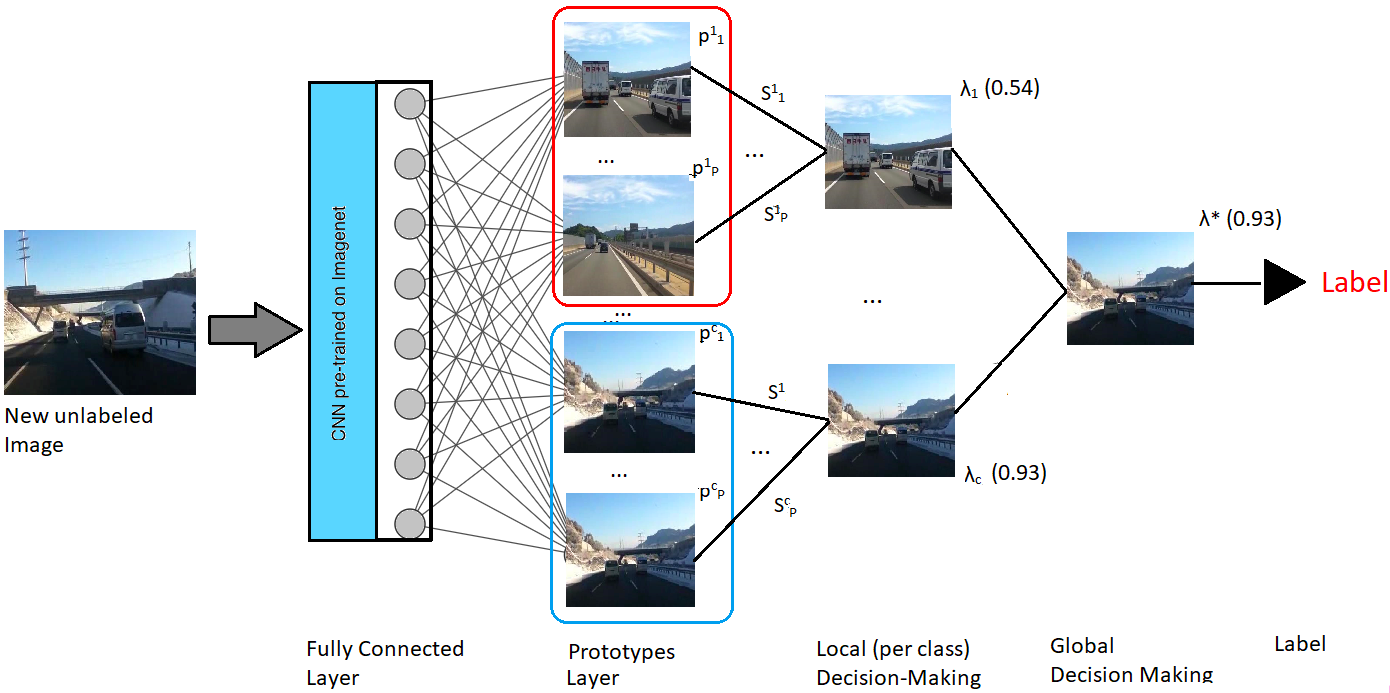}}
		\caption{Architecture for the validation process of the proposed xDNN. } \label{Fig7}
	\end{center}
\end{figure*}

Architecture for the validation process of the proposed xDNN method is illustrated by Figure \ref{Fig7}.

The validation process of xDNN is composed of the following layers: 

\begin{enumerate}
    
    \item Features descriptor layer;
    \item Similarity layer (density);
    \item Local decision-making.
    \item Global decision-making.
    
\end{enumerate}

Which is detailed described as following:
\begin{enumerate}

\item \textbf{Features descriptor layer}:

Similarly to the features descriptor layer described in the training process. 
\item \textbf{Prototypes layer}:

In this layer the degrees of similarity to the nearest prototypes (per class) are extracted for each unlabeled (new/validation) data sample/image $I_i$ defined as follows: 

\begin{eqnarray}
S(x,p_i) = \frac{1}{1+\frac{(x-p_i)}{\sigma^2_i}}, 
\label{eq1}
\end{eqnarray}

\noindent where $S$ denotes the similarity degree.

\item \textbf{Local (per class) decision-making layer}:

Local (per class) decision-making is calculated based on the `winner-takes-all' principle and can be obtained by:
\begin{eqnarray}
\lambda_c = \underset{j=1,2,...,P}{max} (S_j),
\label{eq1}
\end{eqnarray}

\item \textbf{Global decision-making layer}:
The global decision-making layer is in charge of forming the decision by assigning labels to the validation images based on the degree of similarity of the prototypes obtained by the prototype identification layer as illustrated by Figure \ref{Fig7} and determining the winning class.

\begin{eqnarray}
\lambda^*_c = \underset{c=1,2,...,C}{max} (\lambda_c),
\label{eq1}
\end{eqnarray}

In order to determine the overall degree of satisfaction, the maximum of the local, per class winners is applied.

The label is obtained by the following equation (\ref{label}):

\begin{eqnarray}
label = \underset{c=1,2,...,C}{argmax} (\lambda^*_c),
\label{label}
\end{eqnarray}

\end{enumerate}

\section{Experimental Data}

We validated our proposed approach, xDNN using several complex, well-known image classification benchmark datasets (iRoads and Calltech-256).

\subsection{iRoads dataset }

The iROADS dataset \cite{rezaei2013vehicle} was considered in the analysis first. The dataset contains 4,656 image frames recorded from moving vehicles on a diverse set of road scenes, recorded in day, night, under various weather and lighting conditions, as described below: 

\begin{itemize}

\item Daylight - 903 images

\item Night -  1050 images

\item Rainy day - 1049 images

\item Rainy night - 431 images

\item Snowy - 569 images

\item Sun strokes - 307 images

\item Tunnel - 347 images

\end{itemize}

\subsection{Caltech-256}

Caletch-256 has 30,607 images divided into 257 object categories (one of which is the background) \cite{griffin2007caltech}. 

\subsection{Performance Evaluation}

The performance of the classification methods is usually evaluated based on their accuracy index which is defined as follows:

\begin{eqnarray}
ACC(\%) = \frac{TP+TN}{TP+FP+TN+FN},
\label{acc}
\end{eqnarray}

\noindent where $TP, FP, TN, FN$ denote true and false, negative and positive, respectively.

All the experiments were conducted with MATLAB 2018a using a personal computer with a 1.8 GHz Intel Core i5 processor, 8-GB RAM, and MacOS operating system. The classification experiments were executed using 10-fold cross validation under the same ratio of training-to-testing (80\% to 20\%) sample sets.

\section{Results and Analysis}

Computational simulations were performed to assess the accuracy of the proposed explainable deep learning method, xDNN against other state-of-the-art approaches.

\subsection{iRoads Dataset}

\vspace{-5pt}

Table \ref{Table1} shows that the proposed xDNN method provides the best result in terms of classification accuracy as well as time/complexity and simplicity of the model structure (number of parameters/prototypes). The number of model parameters for xDNN (and DRB) is, strictly speaking, zero, because the 2 parameters (mean, $\mu$ and standard deviation, $\sigma$) per prototype (\textit{data cloud}) are derived from the data and are not algorithmic parameters or user-defined parameters. For kNN method one can argue that the number of parameters is the number of data samples, N. The proposed explainable DNN surpasses in terms of accuracy the state-of-the-art VGG--VD--16 algorithm which is a well-established convolutional deep neural network. Moreover, the proposed xDNN has at its top layer a set of a very small number of \textit{MegaClouds} (27 or, on average, 4 \textit{MegaClouds} per class) which makes it very easy to explain and visualize. For comparison, our earlier version of deep rule-based models, called DRB \cite{angelov2018deep} also produced a high accuracy and was trained a bit faster, but ended up with 521 prototypes (on average 75 prototypes per class) \cite{soares2019actively}. With xDNN we do generate meaningful $IF...THEN$ rules as well as generate an analytical description of the \textit{typicality} which is the empirically derived pdf in a closed form which lends itself for further analysis and processing. 

\begin{table}[H]
	\small \caption{Performance Comparasion:  iRoads Dataset}
	\begin{center}
		\begin{tabular}{c|ccc}

			\hline
		    Method &$Accuracy$&Time(s)& \# Parameters \\
			\hline
			xDNN	& \underline{\textbf{99.59\%}} & 4.32 & \underline{\textbf{27}} \\
			VGG--VD--16 \cite{soares2019actively}	& 99.51 \% & 836.28 & Not reported\\
			DRB	\cite{soares2019actively}& 99.02\% & \underline{\textbf{2.95}} & 521 \\
			SVM	\cite{soares2019actively}& 94.17\% & 5.67 & Not reported \\
			KNN	\cite{soares2019actively}& 93.49\% & 4.43 & 4656\\
			Naive Bayes	\cite{soares2019actively}& 88.35\% & 5.31  & Not reported\\
			\hline
		\end{tabular}
		\label{Table1}
	\end{center}
\end{table}

\textit{MegaClouds} generated by the proposed xDNN model can be visualized in terms of rules as illustrated by the Figure 8.

\begin{figure}[h]
	\begin{center}
	IF (I $\sim$ {\includegraphics[scale=0.25]{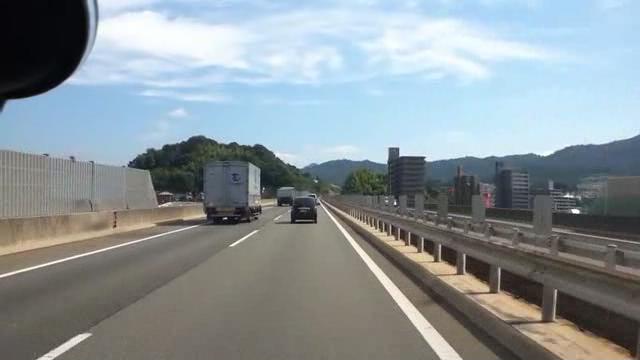}})  OR \nonumber \\ (I $\sim$ {\includegraphics[scale=0.25]{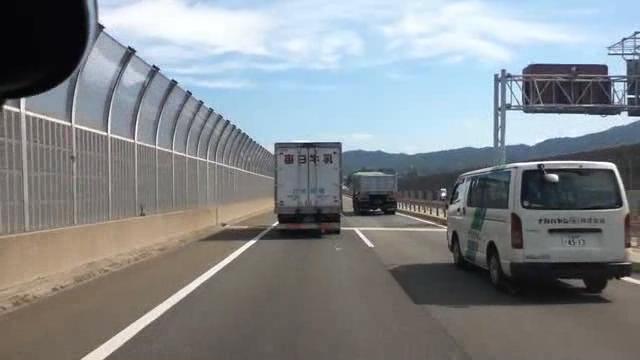}}) OR \nonumber \\ OR (I $\sim$ {\includegraphics[scale=0.25]{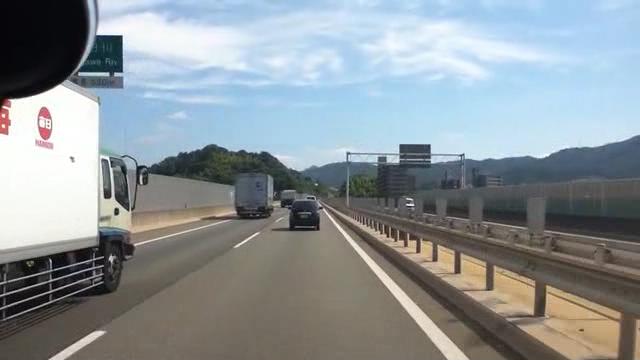}}) \nonumber \\ THEN `Daylight scene'
		\label{xdnnRule}
		\caption{xDNN rule generated for the `Daylight scene'.   }
	\end{center}
\end{figure}

\begin{figure}[h]
	\begin{center}
		{\includegraphics[scale=0.35]{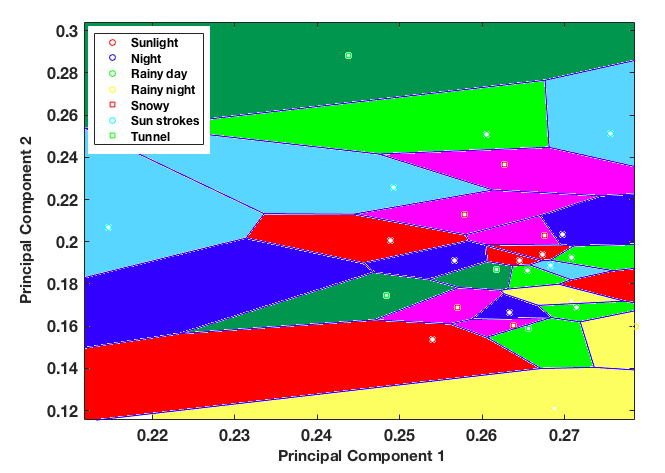}}
		\caption{\textit{MegaClouds} for the iRoads dataset.  } \label{Fig10}
	\end{center}
\end{figure}

Voronoi tesselation can also be used to visualize the resulting \textit{MegaClouds} as illustrated by Figure  \ref{Fig10}.

\begin{figure}[h]
	\begin{center}
		{\includegraphics[scale=0.4]{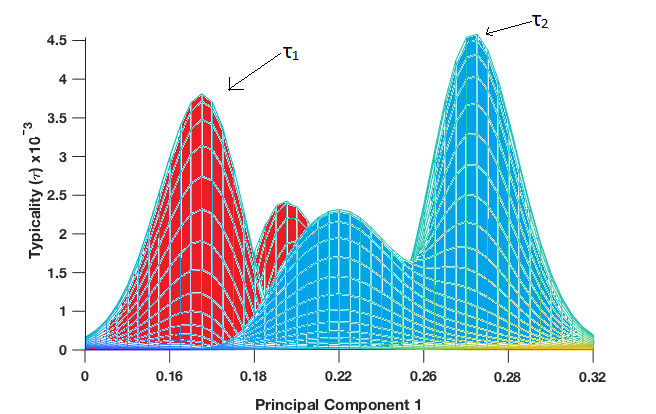}}
		\caption{\textit{Typicality} for the iRoads dataset (2D), 2 classes, representing `night scene' and `snow scene'. } \label{Fig26}
	\end{center}
\end{figure}

\textit{Typicality} for classes `night scene' and `snow scene' are given by Figure \ref{Fig26}.

\textit{Typicality} can also be used for interpreatability and explainability as it is correspondent to the pdf. One can use the \textit{typicality} to represent the likelihood that an image represents a specific type of driving conditions. For a given image a vector of features can be extracted, $x \in R^{4096}$  which can be standardized and normalized and used to demonstrate the likelihood of a certain type of driving condition as shown on Fig. \ref{Fig26}.

\subsection{Caltech-256 Dataset}

Results for Caltech-256 are presented in Table \ref{Table3}. 

\begin{table}[H]
	\small \caption{Performance Comparasion:  Caltech-256 Dataset}
	\begin{center}
		\begin{tabular}{c|c}

			\hline
		    Method &$Accuracy$ \\
			\hline
			xDNN	& \underline{\textbf{75.41}}\% \\
			SVM(1) \cite{zeiler2014visualizing}	& 24.6 \% \\
			SVM(2)	\cite{zeiler2014visualizing}& 39.6\% \\
			SVM(3)	\cite{zeiler2014visualizing}& 46.0\% \\
			SVM(4) \cite{zeiler2014visualizing}& 51.3\% \\
			SVM(5)	\cite{zeiler2014visualizing}& 65.6\%\\
			SVM(7) \cite{zeiler2014visualizing}& 71.7\% \\
			Softmax(5)\cite{zeiler2014visualizing}	& 65.7\% \\
			Softmax(7)	\cite{zeiler2014visualizing}& 74.2\% \\
			\hline
		\end{tabular}
		\label{Table3}
	\end{center}
\end{table}

Results presented in Table \ref{Table3} demonstrate that the proposed xDNN approach can obtain the best classification reported so far world wide for this complex problem, namely, 75.41\%. The proposed approach did surpass all of the competitors, offering the highest accuracy, as well as, clearly explainable model. xDNN produced on average 3 \textit{MegaClouds} per class (a total of 721) which are clearly explainable. Rules have the following format:

\begin{figure}[H]
	\begin{center}
	IF (x $\sim$ {\includegraphics[scale=0.15]{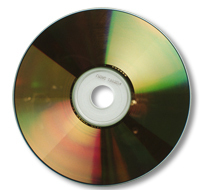}})  OR \nonumber (x $\sim$ {\includegraphics[scale=0.15]{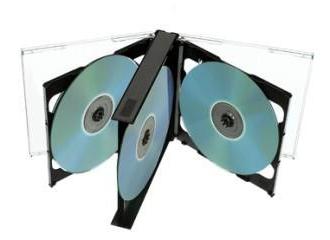}}) OR  (x $\sim$ {\includegraphics[scale=0.15]{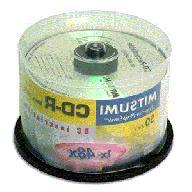}}) \nonumber \\ THEN `CD'
		\label{drb_class}
	\end{center}
\end{figure}

Experiments have demonstrated that the proposed xDNN approach is able to produce highly accurate results surpassing state-of-the-art methods for different challenging datasets. Moreover, xDNN presents highly interpretable results that can be presented in the form of $IF ... THEN$ logical rules, Voronoi tessellations, and/or \textit{typicality} (empirically derived form of pdf) in a closed analytical form allowing further analysis. Because of its recursive, non-iterative and non-parametric form it allows computationally very efficient implementations to be realized.

\section{Conclusion}

In this paper we propose a new method, explainable deep neural network (xDNN), that is directly addressing the bottlenecks of the traditional deep learning approaches and offers a clearly explainable internal architecture that can outperform the existing methods. The proposed xDNN approach requires very little computational resources (no need for GPUs) and short training times (in the order of seconds). The proposed approach, xDNN is prototype-based. Prototypes are actual training data samples (images), which have local peaks of the empirical data distribution called \textit{typicality} as well as of the data density. This generative model is identified in a closed form and equates to the pdf but is derived automatically and entirely from the training data with no user- or problem-specific thresholds, parameters or intervention. The proposed xDNN offers a new deep learning architecture that combines reasoning and learning in a synergy. It is non-iterative and non-parametric, which explains its efficiency in terms of time and computational resources. From the user perspective, the proposed approach is clearly understandable to human users. Results for some well-known benchmark data sets such as iRoads and Caltech-256 show that xDNN outperforms the other methods including state-of-the-art deep learning approaches (VGG--VD--16) in terms of accuracy, time to train and offers a clearly explainable classifier. In fact, the result on the very hard Caltech-256 problem (which has 257 classes) represents a world record \cite{he2015spatial}\footnote{\url{https://martin-thoma.com/sota/}}. Future research will concentrate on the development of a tree-based architecture, synthetic data generation, and local optimization in order to improve the proposed deep explainable approach.  

\bibliography{report}   
\bibliographystyle{IEEEtran}
\end{document}